# Med-DisSeg: A Dispersion-Driven Framework for Fine-Grained Medical Image Segmentation

Zhiquan Chen[1,2], Haitao Wang[1,2], Zhifu Mo[5], Guowei Zou[1,2], Zhen Zhang[4], Xin Li[3], Lijuan Wu[6,*], Hejun Wu[1,2], Weifeng Li[3,*]

*1 School of Computer Science and Engineering, Sun Yat-sen University, Guangzhou, China*

*2 Guangdong Key Laboratory of Big Data Analysis and Processing, Guangzhou, China*

*3 Department of Emergency Medicine, Guangdong Provincial People's Hospital (Guangdong Academy of Medical Sciences), Southern Medical University, Guangzhou, China*

*4 School of Computer Science and Engineering, Huizhou University, Huizhou, China*

*5 Department of Emergency Medicine, Guangdong Cardiovascular Institute, Guangdong Provincial People's Hospital (Guangdong Academy of Medical Sciences), Southern Medical University,Guangzhou, China*

*6 Medical Research Institute, Guangdong Provincial People's Hospital (Guangdong Academy of Medical Sciences), Southern Medical University, Guangzhou, China*

## Abstract

Accurate medical image segmentation is essential for quantitative image analysis and precision medicine. However, the visual similarity between target regions and surrounding tissues leads segmentation models to learn overly compact feature representations, limiting feature separability in the embedding space. This insufficient separability causes segmentation models to generate ambiguous activation maps and imprecise segmentation results. To address this issue, we propose Med-DisSeg, a dispersion-driven medical image segmentation framework that learns more separable feature representations, thereby enabling more accurate reconstruction of fine structures and boundaries. Within Med-DisSeg, a lightweight Dispersive Loss is introduced as an auxiliary regularizer to encourage more dispersed and separable feature representations. Building on these dispersion-regularized representations, the encoder strengthens structure-relevant responses through joint channel and spatial modeling, while the decoder performs multi-scale adaptive calibration to preserve complementary local details and contextual information. Together, these components improve the segmentation of small, deformable, and boundary-ambiguous targets. Extensive experiments on five public datasets covering three imaging modalities demonstrate the effectiveness and robustness of Med-DisSeg across diverse medical image segmentation tasks. The source code and pretrained models will be released upon acceptance.

## Introduction

Medical image segmentation is a fundamental operation in medical image analysis, assigning a semantic label to every pixel so that organs, lesions, and other clinically relevant structures are delineated. It directly enables lesion localization, organ delineation, quantitative assessment, treatment planning, and computer-assisted diagnosis. Accurate segmentation ensures that target regions are characterized precisely, forming a dependable basis for clinical evaluation and precision medicine.

Most existing medical image segmentation methods are built upon the encoder–decoder paradigm established by U-Net (Ronneberger et al. 2015). In this framework, the encoder extracts hierarchical semantic representations, while the decoder progressively restores spatial resolution. Skip connections bridge the two stages by transferring low-level spatial details to the decoder, making the architecture effective for pixel-level prediction. Building on this paradigm, subsequent studies have improved segmentation performance from different perspectives. nnU-Net (Isensee et al. 2021) enhances task adaptability through automated configuration, Attention U-Net (Oktay et al. 2018) introduces attention mechanisms to emphasize target-relevant regions, and Transformer-based models such as TransUNet (Chen et al. 2021) and UNETR (Hatamizadeh et al. 2022) strengthen global context modeling through long-range dependency learning.

Recent approaches, such as ConDSeg (Lei et al. 2025), have applied contrastive representation learning to improve foreground–background separation. However, most existing works rely on predefined contrastive relations or network design changes to sharpen feature discrimination, leaving the global distribution of the learned representations largely unregulated. This becomes especially problematic when target regions and surrounding tissues look similar: heterogeneous anatomical patterns can still be compressed into a feature space that is too narrow to tell them apart. When the learned representations are not sufficiently separable, the encoder becomes less effective at capturing fine-grained structures, which in turn makes the decoder less dependable for reconstructing local details and boundaries. The final outcome is often ambiguous activation maps, incomplete structures, and inaccurate segmentation contours.

To tackle these problems, we introduce Med-DisSeg, a dispersion-driven segmentation framework that learns more separable feature representations and thereby recovers fine structures and boundaries more accurately. A lightweight Dispersive Loss is added as an auxiliary regularizer, pushing in-batch representations to be more dispersed and separable. With these dispersion-regularized features, the encoder enhances structure-related responses with joint channel and spatial modeling, while the decoder uses multi-scale adaptive calibration to keep both local details and contextual information. These design choices jointly improve segmentation on small, deformable, and boundary-ambiguous targets. Unlike contrast-based methods that rely on predefined foreground-background relations, Med-DisSeg directly acts on the distribution of in-batch representations and requires no explicit positive pairs. Our main contributions are as follows:

- We propose Med-DisSeg, a dispersion-driven encoder–decoder framework that leverages these regularized representations for fine-grained segmentation. The encoder enhances structure-relevant responses through joint channel and spatial modeling, while the decoder performs multi-scale adaptive calibration to improve the reconstruction of local details and boundaries.
- We introduce a lightweight Dispersive Loss as an auxiliary regularizer for medical image segmentation, which enlarges inter-sample margins in the embedding space to encourage more dispersed and separable feature representations without additional inference cost.
- Extensive experiments on five public datasets covering three imaging modalities demonstrate the effectiveness and robustness of Med-DisSeg across diverse medical image segmentation tasks.

## Related Works

### Encoder–Decoder Architectures for Medical Image Segmentation

Encoder–decoder networks remain the dominant approach in medical image segmentation. The U-shaped architecture, introduced by U-Net (Ronneberger et al. 2015), has been widely adopted: the encoder captures hierarchical semantic features, while the decoder gradually restores spatial resolution through skip connections. Following this paradigm, nnU-Net (Isensee et al. 2021) showed that careful configuration of preprocessing, network design, and training strategies yields a strong and robust baseline across a variety of segmentation tasks.

Since U-Net, the encoder–decoder framework has been refined along several lines. Attention-based models direct the network toward clinically relevant areas and reduce background interference; boundary-aware and deformable designs, meanwhile, help segment irregular lesions, thin structures, and targets with varying shapes (Oktay et al. 2018; Wang al. 2022; Xu et al. 2023; Jin et al. 2019). Improvements in decoding also allow for better fusion of multi-scale features and more accurate reconstruction of fine details (Rahman et al. 2024). In parallel, transformer-based and hybrid architectures have been introduced to enhance long-range dependency modeling and global context aggregation (Chen et al. 2021; Hatamizadeh et al. 2022; Guo et al. 2025).

Although these architectural advances improve semantic extraction, contextual modeling, and boundary recovery, they mainly enhance network components without explicitly regulating the distribution of learned representations. As a result, when anatomical structures are heterogeneous or target regions closely resemble surrounding tissues, segmentation models tend to learn feature spaces that are too compact. Such representations are difficult to separate, which in turn reduces the reliability of subsequent fine-detail and boundary reconstruction. Med-DisSeg takes a different approach: it regularizes the embedding space directly via a Dispersive Loss, and then makes full use of the resulting separable features through structure-aware encoding and multi-scale decoder calibration.

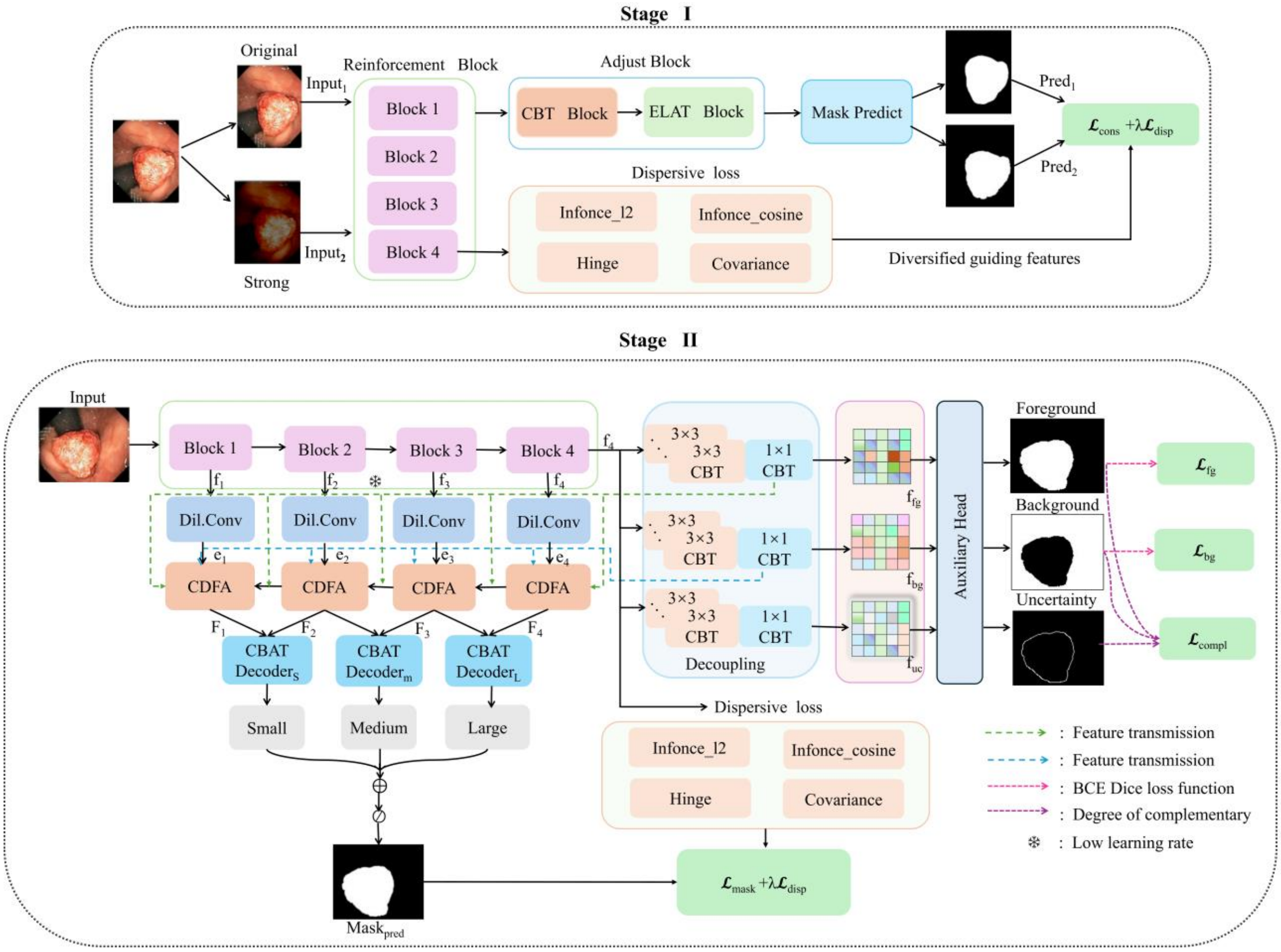

**Figure 1.** Overall architecture of Med-DisSeg: Stage I (top) and Stage II (bottom).

## Representation Regularization for Medical Image Segmentation

Apart from architectural improvements, representation regularization offers another way to enhance medical image segmentation by directly shaping the distribution and separability of learned features. Many contrast-based methods establish foreground–background, region-level, or pixel-level relations to sharpen target discrimination and reduce ambiguous responses (Lei et al. 2025; Ma and Wang 2024; Chen et al. 2024). In addition, representation-level supervision—as explored in SimCLR (Chen et al. 2020), VoCo (Wu et al. 2024), and DyCON (Assefa et al. 2025)—has been shown to benefit downstream medical image analysis.

Although these methods are effective, many of them depend on explicitly constructed positive pairs, carefully designed sampling strategies, auxiliary prediction heads, or additional unlabeled data. Such requirements can complicate the training process and make the resulting representations sensitive to how pairs are defined and how data are organized. Furthermore, predefined foreground–background or region-level relations do not directly control the overall distribution of in-batch representations in the embedding space.

Drawing on the dispersion regularization proposed by Wang and He (2025), Med-DisSeg introduces Dispersive Loss as an auxiliary regularizer for medical image segmentation, with the goal of improving feature separability. Unlike contrast-driven methods that rely on predefined semantic relations, Med-DisSeg directly regularizes the distribution of in-batch representations and then leverages the resulting separable features through structure-relevant encoding and multi-scale decoder calibration.

## Methodology

Given an input medical image $I \in \mathbb{R}^{H\times W\times C}$, Med-DisSeg predicts a pixel-wise segmentation map $\hat{Y} \in \mathbb{R}^{H\times W\times K}$, where $C$ is the number of input channels and $K$ denotes the number of target classes. For binary segmentation, $K = 1$ and the output represents the foreground probability map. The framework follows a two-stage training pipeline. In Stage I, the encoder is regularized to learn robust and separable representations. In Stage II, the regularized encoder is coupled with a multi-scale decoder for fine-grained mask prediction. We then detail the Dispersive Loss, the encoder-side adaptive attention mechanism, and the decoder-side multi-scale calibration strategy. Together, these components improve feature-space separability, refine target- and boundary-relevant responses, and enhance the reconstruction of small and ambiguous regions.

## Network Architecture

Med-DisSeg adopts a two-stage framework that separates dispersion-regularized encoder learning from multi-scale mask decoding, as illustrated in Fig. 1. Stage I learns a robust encoder with auxiliary supervision, while Stage II integrates the pre-trained encoder into the full encoder–decoder network for fine-grained segmentation.

**Stage I: Encoder pre-training.** we pre-train the encoder using a ResNet-50 backbone. Strong photometric perturbations are applied during training to improve robustness against appearance variations. The encoder is augmented with the Encoder-side Local Adaptive TeLU attention (ELAT) module, and a lightweight prediction head is attached on top, supervised jointly by the segmentation loss and the Dispersive Loss. Since this prediction head is removed after pre-training, the performance gains observed at this stage can be attributed primarily to the dispersion-regularized encoder representations, rather than to any additional decoding capacity.

**Stage II: Multi-scale feature decoding.** In Stage II, the pre-trained encoder is incorporated into the full Med-DisSeg framework and fine-tuned with a smaller encoder learning rate, while Dispersive Loss is retained as an auxiliary regularizer. Given an input image, the encoder produces hierarchical feature maps $\{f_1, \dots, f_4\}$. The deepest feature $f_4$ is further processed to generate foreground, background, and uncertainty-aware guidance features, denoted as $f_{\mathrm{fg}}$, $f_{\mathrm{bg}}$, and $f_{\mathrm{uc}}$. The corresponding branches are used for feature guidance rather than as independent prediction objectives; the main optimization objective consists of the mask loss and Dispersive Loss. These guidance features are used by the Complementary Decoupled Feature Aggregation (CDFA) module to aggregate multi-level features and suppress distracting background responses. Finally, three Channel-Balanced Adaptive TeLU attention (CBAT) decoders with different receptive fields decode complementary small-, medium-, and large-scale cues, and their outputs are fused to obtain the final segmentation mask.

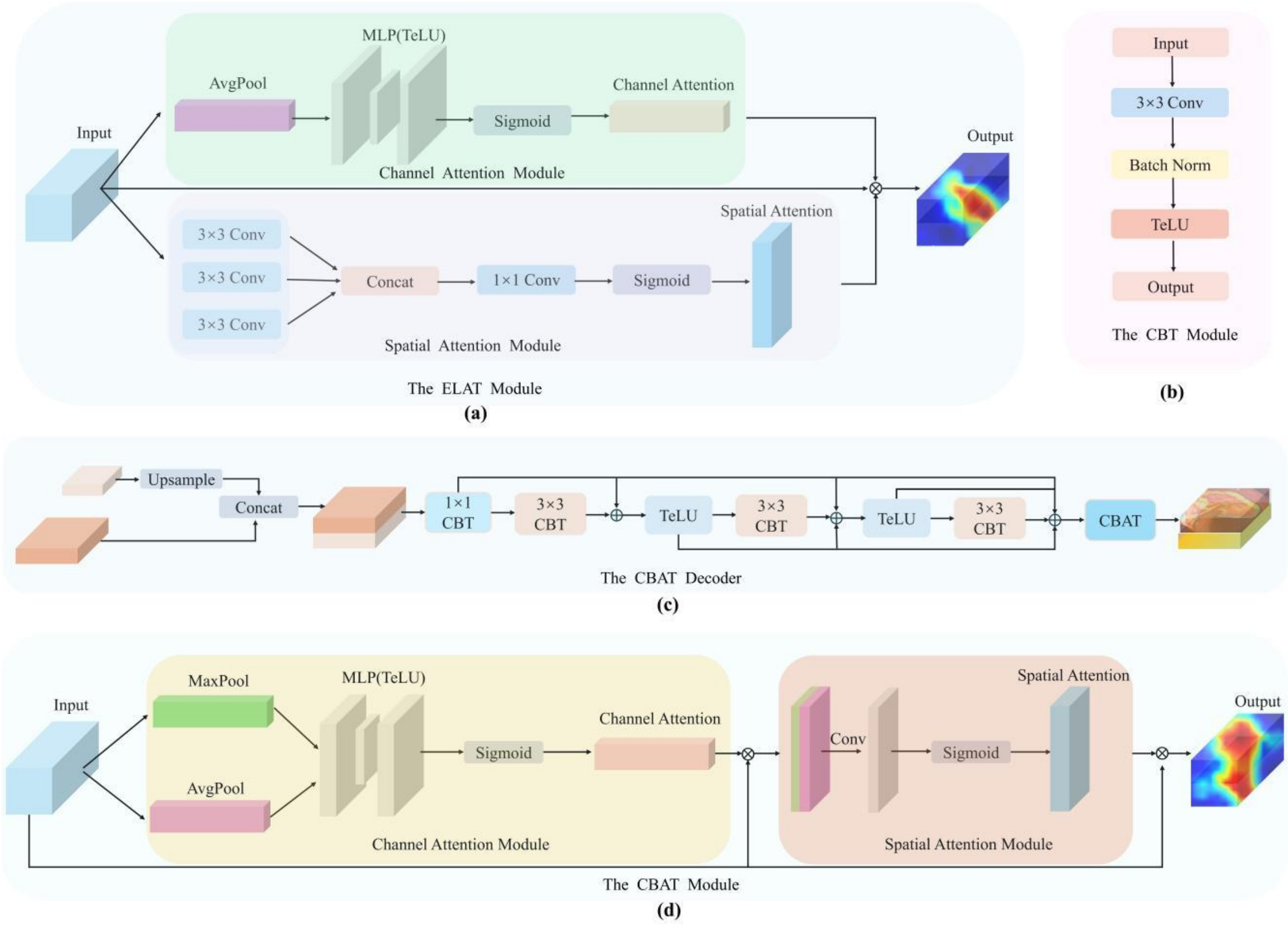

**Figure 2.** (a) Architecture of the ELAT module, (b) structure of the CBT blocks, (c) overall design of the CBAT-based decoder, and (d) internal structure of the CBAT module.

## Improving Feature Separability with Dispersive Loss

Under limited supervision and heterogeneous appearances, medical image segmentation models can end up with overly compact feature distributions, making it harder to discriminate between visually similar regions. This excessive concentration reduces feature separability and often leads to ambiguous activations, boundary leakage, and false background responses. To address this, we introduce Dispersive Loss as a lightweight auxiliary regularizer in both Stage I and Stage II. By treating in-batch hidden representations as mutually negative samples, the loss enlarges inter-sample margins in the embedding space, yielding more separable features that benefit the subsequent attention-enhanced encoding and multi-scale decoding.

The key idea of Dispersive Loss is to obtain a "contrastive-style objective without explicit positive pairs". Starting from a simplified InfoNCE objective (Wang and Liu 2021),

$$\mathcal{L}_{\text{InfoNCE}} = \frac{\mathcal{D}(z_i, z_i^+)}{\tau} + \log \sum_j \exp\left(-\frac{\mathcal{D}(z_i, z_j)}{\tau}\right). \tag{1}$$

where $z_i$ and $z_i^+$ form a positive pair, $z_j$ are negatives, $\mathcal{D}(\cdot,\cdot)$ is a distance (or negative similarity), and $\tau$ is the temperature. The first term enforces positive alignment, and the second encourages dispersion among all samples. Dispersive Loss discards the alignment term and keeps only the dispersion term:

$$\mathcal{L}_{\text{Disp}} = \log \mathbb{E}_{i \neq j}\left[\exp\left(-\frac{\mathcal{D}(z_i, z_j)}{\tau}\right)\right]. \tag{2}$$

This objective encourages greater dispersion among in-batch representations and improves feature-space separability without requiring explicit positive pairs.

In Med-DisSeg, we instantiate four variants of Dispersive Loss by removing the positive-pair alignment term from different contrastive or regularization objectives. Let $\{h_i\}_{i=1}^{B}$ be the batch representations, $\tau > 0$ the temperature, and $\varepsilon > 0$ a small numerical stabilizer. Self-pairs are excluded ($i \neq j$).

**1. InfoNCE-based Dispersive Loss with L2 distance:**

$$\mathcal{L}_{\text{disp}}^{\text{InfoNCE-L2}} = \log\left(\mathbb{E}_{i \neq j}\left[\exp\left(-\| h_i - \frac{h_j \|_2^2}{\tau}\right)\right] + \varepsilon\right). \tag{3}$$

Here $D(h_i, h_j) = \| h_i - h_j \|_2^2$ measures Euclidean discrepancy in the representation space. This formulation encourages geometric separation among features, while $\varepsilon > 0$ ensures numerical stability.

**2. InfoNCE-based Dispersive Loss with cosine distance:**

$$\mathcal{L}_{\text{disp}}^{\text{InfoNCE-Cos}} = \log\left(\mathbb{E}_{i \neq j}\left[\exp\left(-\frac{1 - \hat{h}_i^\top \hat{h}_j}{\tau}\right)\right] + \varepsilon\right), \quad \hat{h}_i = \frac{h_i}{\|h_i\|_2}. \tag{4}$$

This variant relies on cosine dissimilarity $D(h_i, h_j) = 1 - h_i^\top h_j / \|h_i\|_2 \left\|h_j\right\|_2$, which captures angular differences between normalized representations. It is scale-invariant and focuses on directional diversity.

**3. Hinge-based Dispersive Loss:**

$$\mathcal{L}_{\text{disp}}^{\text{Hinge}} = \mathbb{E}_{i \neq j}\left[\max(0,\ \delta - \left\|h_i - h_j\right\|_2)^2\right]. \tag{5}$$

This hinge-style formulation enforces a minimum margin $\delta$ between any pair of representations. It directly penalizes pairs that are closer than the margin, providing explicit control over the minimum dispersion distance.

**4. Covariance off-diagonal penalty:**

$$\mathcal{L}_{\text{disp}}^{\text{Cov}} \& = \sum_{p \neq q} = C_{pq}^2 \|C - \text{diag}(C)\|_F^2, \\ C = \frac{1}{B-1} \widetilde{H}^\top \widetilde{H}, \qquad \widetilde{H} = H - \mathbf{1}\bar{h}^\top. \tag{6}$$

where $H = [h_1^\top, \dots, h_B^\top]^\top$ and $\bar{h} = 1/B \sum_i h_i$. This variant minimizes the off-diagonal energy of the covariance matrix computed from centered features. By discouraging cross-dimensional correlations, it promotes feature decorrelation and reduces redundancy across representation dimensions.

**Overall objective.** In both Stage I and Stage II, Dispersive Loss serves as a generic regularizer that encourages feature dispersion and improves representation separability, while the segmentation head and architectural modules specialize these features for mask prediction. The total loss in both stages is defined as

$$\mathcal{L}_{\text{total}} = \mathcal{L}_{\text{mask}} + \lambda \mathcal{L}_{\text{disp}}. \tag{7}$$

where $\mathcal{L}_{\text{mask}}$ denotes the segmentation loss, $\mathcal{L}_{\text{disp}}$ the chosen Dispersive Loss variant, and $\lambda$ is a balancing hyperparameter. In the next subsection, we describe how ELAT and CBAT are designed to exploit these dispersion-regularized features for fine-grained and multi-scale decoding.

## Attention-enhanced feature encoding and multi-scale decoding

Dispersive Loss contributes to feature-space separability, but accurate segmentation additionally requires the network to highlight target-relevant regions and reconstruct details at the right scales. Med-DisSeg therefore incorporates two further components on top of the regularizer: an encoder-side adaptive attention mechanism, which refines target- and boundary-relevant encoder responses, and a decoder-side multi-scale calibration strategy, which combines local detail with contextual information for more precise mask reconstruction.

## Encoder-side adaptive attention

In medical images, targets often exhibit weak contrast and ambiguous boundaries, making it difficult for the encoder to distinguish relevant regions from surrounding tissues. To improve feature refinement, we introduce an encoder-side adaptive attention mechanism, as shown in Fig. 2(a). Given an input feature map $\mathrm{X} \in \mathbb{R}^{\mathrm{B \times C \times H \times W}}$, the ELAT module computes channel and spatial attention in parallel:

$$\mathrm{A_c} = \sigma\big(\mathrm{MLP_{TeLU}}(\mathrm{GAP}(\mathrm{X}))\big), \\ X_s = [\mathrm{Conv}_{3\times3}^{r=1}(X), \mathrm{Conv}_{3\times3}^{r=2}(X), \mathrm{Conv}_{3\times3}^{r=3}(X)], \\ A_s = \sigma(\mathrm{Conv}_{1\times1}^{3C\to1}(X_s)).$$

$$Y = X \oplus (X \odot A_c \odot A_s). \tag{8}$$

Here, $A_c \in \mathbb{R}^{B\times C\times 1\times 1}$ models channel importance, $A_s \in \mathbb{R}^{B\times 1\times H\times W}$ models spatial saliency. $GAP(\cdot)$ denotes the global average pooling operation, $Conv^{r}_{3\times 3}$ represents a $3 \times 3$ convolution with a dilation rate of r, $Conv^{3C\to 1}_{1\times 1}$ is a $1 \times 1$ convolution mapping $3C$ channels to a single channel, $\sigma(\cdot)$ denotes the sigmoid function, $[\cdot]$ denotes channel-wise concatenation, and $\odot$ and $\oplus$ represent element-wise multiplication and addition with broadcasting, respectively.

**Decoder-side multi-scale adaptive calibration**

Although the encoder provides refined features, small lesions, thin structures, and deformable regions may still be weakened during decoding. To recover such regions more effectively, we design a multi-scale calibration strategy with three decoding paths, denoted as $D_s$, $D_m$, and $D_l$, corresponding to small, medium, and large receptive fields. These paths capture fine local details, intermediate semantic cues, and broader contextual information, respectively, and thus provide complementary evidence for final mask prediction.

As shown in Fig. 2(c), each decoding path takes two input features with different spatial resolutions, denoted by $F_1$ and $F_2$. The higher-level feature $F_1$ is first upsampled and concatenated with $F_2$. The concatenated feature is then refined by stacked CBT blocks with residual shortcuts and calibrated by the CBAT unit:

$$\begin{aligned} F_\ell^{(0)} &= [\,\mathrm{Up}(F_1);\, F_2\,], \\ F_\ell^{(t+1)} &= F_\ell^{(t)} + \mathrm{CBT}_{3\times 3}\left(\mathrm{CBT}_{1\times 1}(F_\ell^{(t)})\right), \quad t = 0,1,2, \\ O_\ell &= \mathrm{CBAT}\left(F_\ell^{(3)}\right), \quad \ell \in \{S, M, L\}. \end{aligned} \tag{9}$$

where $\mathrm{Up}(\cdot)$ denotes upsampling, $\mathrm{CBT}(\cdot)$ denotes the convolution–batch normalization–TeLU block, and $\mathrm{CBAT}(\cdot)$ denotes the adaptive calibration unit.

The final prediction is obtained by aggregating the three scale-specific outputs:

$$O = O_S + O_M + O_L. \tag{10}$$

This aggregation combines multi-scale cues and helps preserve both target details and boundary information in the final segmentation mask.

## Experiments

### Experimental Setup

All experiments were run on a single NVIDIA GeForce RTX 4090 GPU. The input resolution was fixed to $256 \times 256$ pixels. We applied lightweight data augmentation comprising random rotations and vertical/horizontal flips. Training used mini-batches of size 4 and the Adam optimizer (Kingma and Ba 2015). Unless otherwise stated, ResNet-50 (He et al. 2016) served as the default encoder. We adopted a two-stage schedule: in Stage I the learning rate was $1 \times 10^{-4}$ for the whole network; in Stage II we initialized from the Stage I encoder and reduced the encoder's learning rate to $1 \times 10^{-5}$ while keeping $1 \times 10^{-4}$ for the remaining parts. Throughout all experiments, we reproduce competing methods using their official implementations whenever available; when official code is unavailable, we report the results from the original papers and clearly indicate this in the corresponding tables/captions.

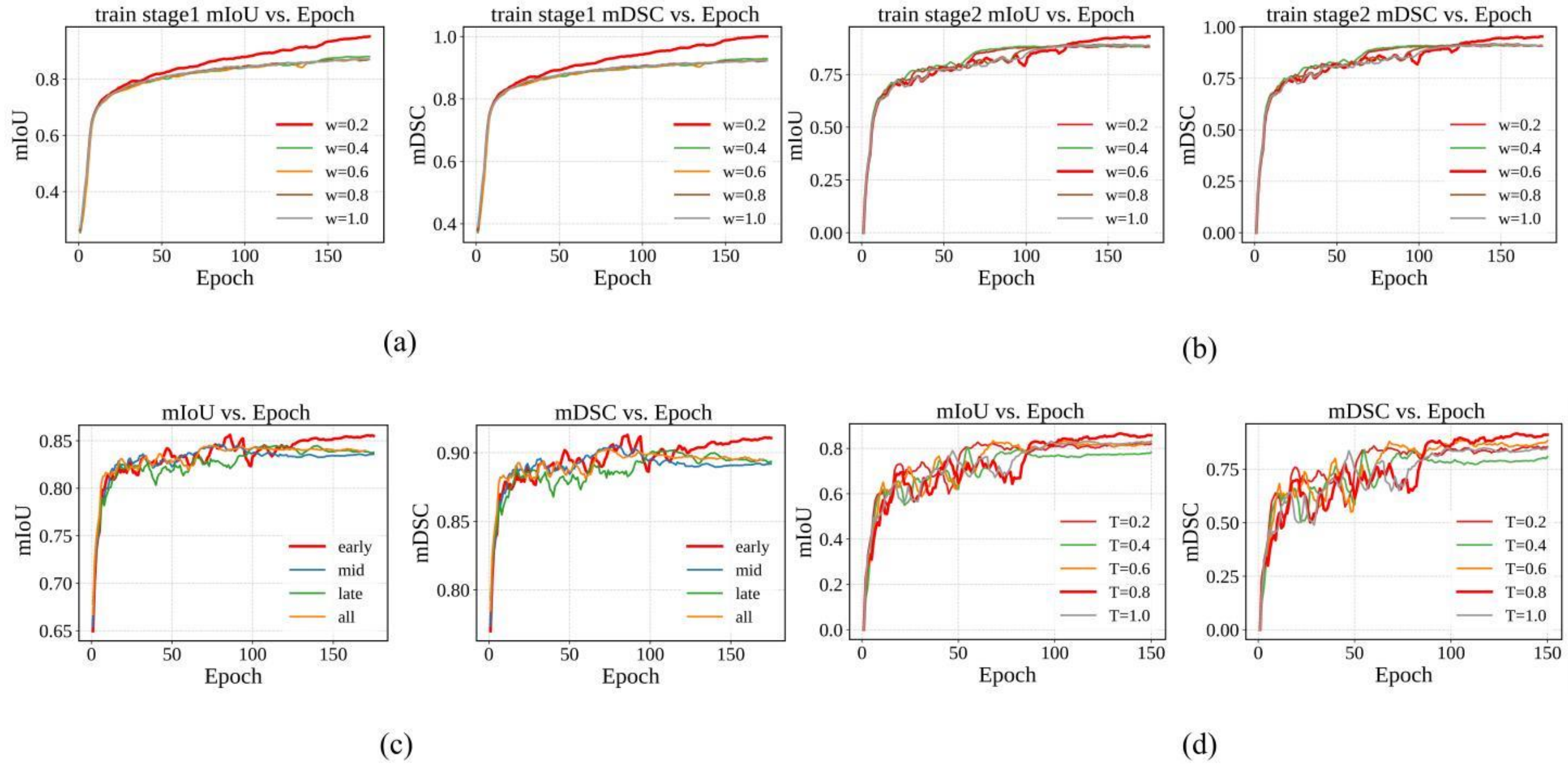

**Figure 3.** Hyperparameter analysis of the InfoNCE–L2 Dispersive Loss. (a) and (b) show the effects of the loss weights in Stage I and Stage II, respectively. (c) evaluates different encoder-layer placements for applying the regularizer. (d) analyzes the influence of the temperature parameter.

## Datasets and Metrics

Our evaluation relies on five widely used segmentation benchmarks: Kvasir-SEG (Jha et al. 2020), Kvasir-Sessile (Jha et al. 2021), GlaS (Sirinukunwattana et al. 2017), ISIC-2016 (Gutman et al. 2016), and ISIC-2017 (Codella et al. 2019). The chosen datasets cover three imaging modalities and diverse tasks, from polyp and gland segmentation to skin lesion delineation. More importantly, they differ noticeably in object size, shape, contrast, and boundary ambiguity, which allows us to examine how well Med-DisSeg handles challenging fine-grained cases.

All datasets are split according to official or widely accepted protocols. Kvasir-SEG uses 880 training and 120 validation samples; Kvasir-Sessile is split into 156/20/20 for training/validation/testing (Tomar et al. 2022; Jha et al. 2021). GlaS, ISIC-2016, and ISIC-2017 are evaluated with their official splits: 85/80, 900/379, and 2000/150/600, respectively. In each case, the foreground class refers to the target structure—polyps, glands, or skin lesions—and all remaining pixels are assigned to the background.

For evaluation, we use five metrics: mean Intersection-over-Union (mIoU), mean Dice Similarity Coefficient (mDSC), Recall, Precision, and Boundary F1 (BF1). mIoU and mDSC are the main overlap-based measures; Recall and Precision reflect sensitivity to targets and over-segmentation tendency, respectively. BF1, computed with a 2-pixel tolerance, assesses boundary alignment. All metrics are calculated per image, then averaged over the evaluation set, and reported as percentages by default.

**Table 1.** Comparison with other methods on the Kvasir-Sessile, Kvasir-SEG, and GlaS datasets. Bold indicates the best performance and underline indicates the second-best. Results marked with * are reported from the original papers, while unmarked baseline results are reproduced under our experimental settings.

| Methods | KS mIoU | KS mDSC | KS Rec. | KS Prec. | KSEG mIoU | KSEG mDSC | KSEG Rec. | KSEG Prec. | GlaS mIoU | GlaS mDSC | GlaS Rec. | GlaS Prec. |
|---|---|---|---|---|---|---|---|---|---|---|---|---|
| U-Net (MICCAI 2015) | 23.1 | 33.8 | 45.1 | 46.6 | 65.5 | 75.8 | 83.6 | 77.6 | 75.8 | 85.5 | 90.3 | 82.8 |
| U-Net++ (MICCAI 2018) | 38.4 | 50.2 | 62.5 | 51.8 | 67.9 | 77.2 | 86.5 | 77.7 | 77.6 | 86.9 | 89.6 | 85.5 |
| Attn U-Net (MIDL 2018)* | 27.5 | 38.9 | 59.0 | 44.4 | 67.6 | 77.4 | 83.9 | 79.9 | 76.6 | 85.9 | 91.8 | 82.2 |
| PraNet (MICCAI 2020)* | 66.7 | 77.4 | 80.7 | 82.4 | 83.0 | 89.4 | 90.6 | 91.3 | 71.8 | 83.0 | 90.9 | 78.0 |
| TGANet (MICCAI 2022)* | 74.4 | 82.0 | 79.3 | 85.9 | 83.3 | 89.2 | 91.3 | 91.2 | 77.1 | 81.8 | 84.7 | 80.2 |
| DCSAU-Net (Comput. Biol. Med. 2023)* | 72.6 | 81.1 | 65.6 | 62.9 | 83.5 | 88.9 | 89.5 | 89.5 | 77.6 | 86.5 | 93.0 | 82.5 |
| XBFormer (TMI 2023)* | 73.6 | 81.1 | 87.2 | 76.3 | 83.8 | 89.0 | 89.8 | 87.2 | 73.7 | 84.3 | 84.0 | 85.7 |
| CASF-Net (Comput. Biol. Med. 2023)* | 60.5 | 72.4 | 78.0 | 74.8 | 81.7 | 88.7 | 89.2 | 88.2 | 78.4 | 87.2 | 91.3 | 85.9 |
| DTAN (KBS 2024)* | 76.4 | 84.2 | 84.2 | 85.9 | 84.1 | 90.4 | 91.6 | 92.0 | 78.5 | 87.9 | 88.5 | 90.2 |
| DoubleAANet (Information Fusion 2025)* | 72.2 | 82.3 | 89.9 | 79.4 | 82.1 | 90.4 | 91.3 | 76.3 | 83.2 | 85.6 | 88.1 | 90.1 |
| ConDSeg (AAAI 2025) | 81.2 | 89.1 | 90.1 | 90.0 | 84.6 | 90.5 | **92.3** | 91.7 | 85.1 | 91.6 | 93.5 | 90.5 |
| **Ours** | **84.6** | **91.3** | **90.4** | **93.3** | **85.9** | **91.6** | 91.3 | **94.0** | **85.7** | **92.2** | **93.7** | **91.4** |

**Table 2.** Comparison on ISIC-2016 and ISIC-2017 in terms of mIoU and mDSC.

| Method | ISIC-2016 mIoU | ISIC-2016 mDSC | ISIC-2017 mIoU | ISIC-2017 mDSC |
|---|---|---|---|---|
| U-Net (MICCAI 2015) | 83.6 | 90.3 | 73.7 | 82.8 |
| CE-Net (TMI 2019)* | 84.6 | 90.9 | 76.4 | 84.8 |
| CPFNet (TMI 2020)* | 84.2 | 90.7 | 76.2 | 84.7 |
| FAT-Net (Med. Image Anal. 2022)* | 85.3 | 91.6 | 76.5 | 85.0 |
| DCSAU-Net (Comput. Biol. Med. 2023)* | 85.3 | 91.4 | 76.1 | 85.0 |
| EIU-Net (Comput. Biol. Med. 2023)* | 85.5 | 91.9 | 77.1 | 85.5 |
| ADZUS (Bioengineering 2025)* | 86.7 | 92.9 | 80.0 | 88.7 |
| ConDSeg (AAAI 2025) | 86.8 | 92.5 | 80.9 | 88.3 |
| **Ours** | **87.4** | **93.1** | **81.4** | **89.7** |

## Comparison with Other State-of-the-Art

Tables 1 and 2 report the main overlap-based results, and Table 3 reports the boundary-aware results. We compare Med-DisSeg with classical encoder–decoder CNNs (Ronneberger et al. 2015; Zhou et al. 2018; Oktay et al. 2018; Gu et al. 2019; Feng et al. 2020), boundary- or context-enhanced models (Fan et al. 2020; Wu et al. 2022; Tomar et al. 2022; Xu et al. 2023), Transformer-based and hybrid architectures (Wang et al. 2023; Zheng et al. 2023; Yu et al. 2023; Zhao et al. 2024), and the contrast-driven method ConDSeg (Lei et al. 2025). These baselines cover diverse architectural designs and representation learning strategies, providing a comprehensive benchmark for evaluating Med-DisSeg.

Med-DisSeg achieves the best overall results on all five datasets under the adopted protocol. Compared to the strong contrast-driven baseline ConDSeg (Lei et al. 2025), our method raises mIoU by up to 3.4%, with an average gain of about 1.3%. The improvements are especially noticeable on difficult cases with small targets and unclear boundaries. This indicates that the combination of dispersion-driven representation learning and attention-enhanced multi-scale decoding helps separate features more cleanly while retaining target-relevant details.

**Table 3.** Boundary-aware comparison on representative datasets.

| Method | Kvasir-SEG BF1 | Kvasir-Sessile BF1 | ISIC-2016 BF1 |
|---|---|---|---|
| U-Net | 74.3 | 41.6 | 84.2 |
| PraNet | 85.1 | 74.8 | 85.6 |
| DCSAU-Net | 86.4 | 78.2 | 87.5 |
| ConDSeg | <u>88.7</u> | <u>85.4</u> | <u>90.3</u> |
| **Ours** | **90.6** | **88.9** | **92.1** |

To further evaluate boundary delineation quality, Table 3 reports BF1 scores on representative datasets. Med-DisSeg consistently achieves the best BF1 results, indicating better boundary alignment than competing methods and further supporting its effectiveness in boundary-ambiguous segmentation scenarios.

## Ablation Study

We conducted extensive ablation studies on the proposed components, and the results are summarized as follows:

**Evaluation of dispersive loss variants.** We compare four dispersive-loss variants on Kvasir-SEG: *InfoNCE–L2*, *InfoNCE–Cosine*, *Hinge*, and *Covariance*. All variants are evaluated with the same loss weight, temperature, and layer placement. As shown in Fig. 4(a), *InfoNCE–L2* performs best, achieving 85.6 mIoU and 91.2 mDSC. Its advantage over *Hinge*, *InfoNCE–Cosine*, and *Covariance* suggests that squared $\ell_2$ distance provides more effective geometric separation in the embedding space. Therefore, *InfoNCE–L2* is adopted as the default dispersive regularizer.

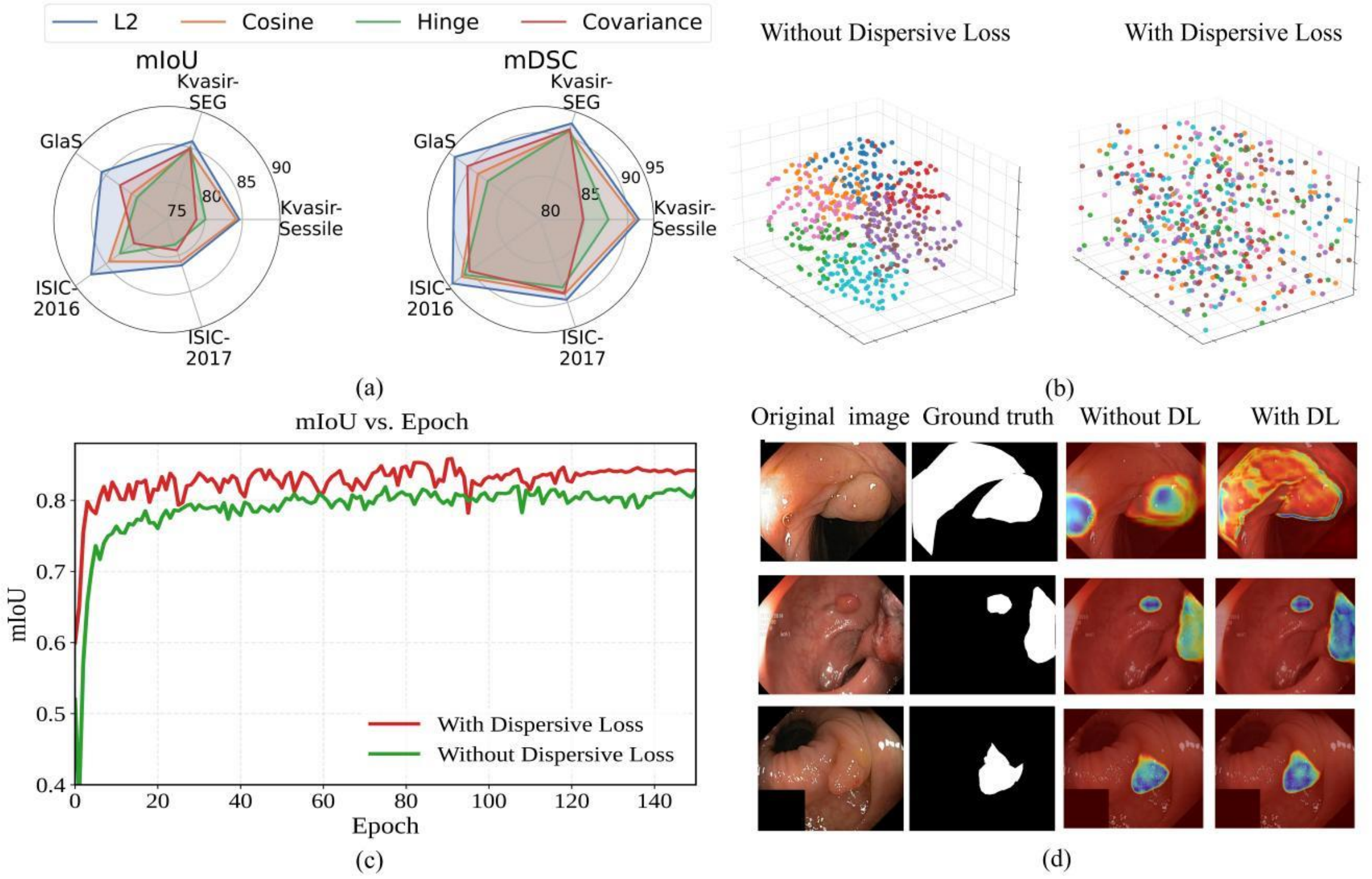

**Figure 4.** Analysis of Dispersive Loss. (a) Comparison of four Dispersive Loss variants across five datasets. (b) PCA visualization of feature embeddings with and without Dispersive Loss. (c) Training curves with and without Dispersive Loss. (d) Grad-CAM maps with and without Dispersive Loss.

**Dispersive Loss hyperparameter study.** We study the effects of the loss weights, layer placement, and temperature on Kvasir-SEG. As shown in Fig. 3, the optimal weights are $\lambda_1 = 0.2$ for Stage I and $\lambda_2 = 0.6$ for Stage II, indicating that mild dispersion is sufficient for encoder pre-training, while stronger regularization benefits fine-tuning. Applying Dispersive Loss at early encoder layers outperforms middle, late, and all-layer settings, suggesting that early feature-space regularization better supports downstream decoding. The best temperature is $\tau = 0.8$, which provides a suitable balance between sharpness and stability. Accordingly, we adopt $(\lambda_1, \lambda_2) = (0.2, 0.6)$, early placement, and $\tau = 0.8$ in subsequent experiments.

**Table 4.** Evaluation on Kvasir-SEG of the ELAT module, CBAT decoder, and Dispersive Loss applied in two stages.

| Variant | Backbone | ELAT | CBAT | Disp-I | Disp-II | mIoU | mDSC |
|---|---|---|---|---|---|---|---|
| 1 | Yes | No | No | No | No | 84.3 | 89.7 |
| 2 | Yes | Yes | No | No | No | 85.0 | 90.8 |
| 3 | Yes | No | Yes | No | No | 84.6 | 90.2 |
| 4 | Yes | Yes | Yes | No | No | 85.2 | 91.1 |
| 5 | Yes | Yes | Yes | Yes | No | 85.4 | 91.3 |
| 6 | Yes | Yes | Yes | No | Yes | 85.6 | 91.4 |
| 7 | Yes | Yes | Yes | Yes | Yes | **85.9** | **91.6** |

**Contribution analysis of the proposed design.** Table 4 reports the ablation results on Kvasir-SEG. Adding either the encoder-side attention or the multi-scale decoder calibration alone improves the baseline, and combining them leads to further gains, which suggests that the two modules complement each other in feature refinement and fine-grained decoding. The Dispersive Loss also brings additional improvements when applied in either training stage, and the best performance is achieved when it is used in both Stage I and Stage II. Overall, these results indicate that dispersion regularization, attention-enhanced encoding, and multi-scale decoder calibration all contribute to more separable representations and stronger segmentation performance.

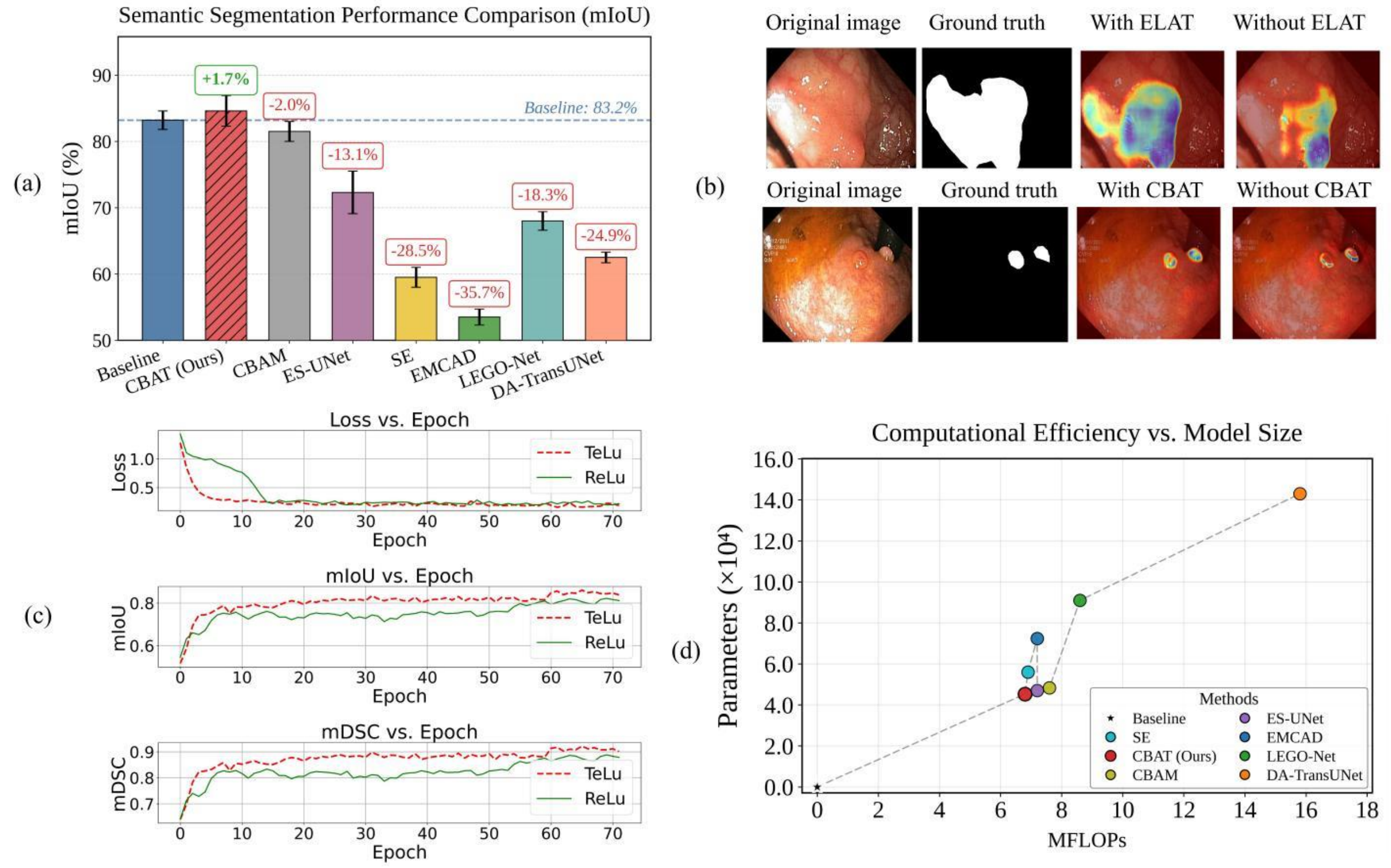

**Figure 5.** We analyze the proposed attention and activation designs from four perspectives. (a) Benchmarks CBAT against typical decoder-side attention mechanisms on Kvasir-SEG. (b) Visualizes the effects of ELAT and CBAT through paired response maps. (c) Compares TeLU with ReLU regarding convergence speed and segmentation accuracy. (d) Summarizes the accuracy–efficiency trade-off across attention variants.

**Comparison of decoder-side attention designs.** To see how CBAT performs relative to other lightweight attention modules, we ran comparisons on Kvasir-SEG using SE, CBAM, EMCAD, and a few recent attention designs (Hu et al. 2018; Woo et al. 2018; Rahman et al. 2024; Park et al. 2025; Xu et al. 2025; Sun et al. 2024). Every variant was placed at the same spot in the decoder and tested under identical conditions, with the final numbers averaged over three random seeds.

As shown in Fig. 5(a), CBAT achieves the best mean mIoU, improving the baseline by 1.7%. Fig. 5(d) further shows its favorable balance between accuracy and efficiency.

**Activation-function validation.** We compare TeLU, defined as $\phi_{\mathrm{TeLU}}(t) = t\tanh(\exp t)$, with ReLU in the encoder-side attention and decoder-side calibration modules under identical settings. As shown in Fig. 5(c), TeLU achieves faster convergence, smoother training curves, and higher validation mIoU/mDSC. Given its negligible overhead, TeLU is adopted as the default activation in Med-DisSeg.

## Generalization Experiment

To assess cross-task generalization, we evaluate Med-DisSeg on the Synapse multi-organ CT dataset. Following the settings of TransUNet and SwinUNet, we convert each 3D CT volume into a sequence of 2D slices and replace the binary loss with a multi-class cross-entropy loss for segmenting eight organs. Compared with the binary lesion datasets used in the main experiments, Synapse presents more diverse organ sizes, shapes, and boundary characteristics.

**Table 5.** Organ-wise performance of different methods on the Synapse dataset reported as DSC (%).

| Organ | UNet | TransUNet | TransNorm* | MT-UNet* | SwinUNet* | ConDSeg | WMREN | Ours |
|---|---|---|---|---|---|---|---|---|
| Aorta | 89.1 | 87.2 | 86.2 | 87.9 | 85.5 | 86.6 | 88.9 | 87.6 |
| Gallbladder | 69.7 | 63.1 | 65.1 | 65.0 | 66.5 | 72.5 | 74.6 | 76.2 |
| Left kidney | 77.8 | 81.9 | 82.2 | 81.5 | 83.3 | 85.1 | 88.5 | 81.6 |
| Right kidney | 68.6 | 77.0 | 78.6 | 77.3 | 79.6 | 78.6 | 84.2 | 84.0 |
| Liver | 93.4 | 94.1 | 94.2 | 93.1 | 94.3 | 90.5 | 95.1 | 94.3 |
| Pancreas | 54.0 | 55.9 | 55.3 | 59.5 | 56.6 | 61.2 | 69.6 | 72.4 |
| Spleen | 86.7 | 85.1 | 89.5 | 87.8 | 90.7 | 87.3 | 91.1 | 91.0 |
| Stomach | 75.6 | 75.6 | 76.0 | 76.8 | 76.6 | 79.8 | 83.1 | 80.1 |
| **Mean DSC** | 76.9 | 77.5 | 78.4 | 78.6 | 79.1 | 80.2 | **84.4** | 83.4 |

As shown in Table 5, Med-DisSeg achieves a mean DSC of 83.4%, outperforming several CNN- and Transformer-based baselines and ranking second overall. While its average score is slightly lower than WMREN (Lu et al. 2025), a method specialized for Synapse, Med-DisSeg obtains the highest DSC on the gallbladder and pancreas. These results suggest that Med-DisSeg can be extended to multi-class CT segmentation and remains effective for small, deformable, and boundary-ambiguous targets.

## Visualization Analysis

Fig. 6 presents qualitative comparisons on Kvasir-SEG and ISIC-2016. Compared with representative baselines, Med-DisSeg produces more complete lesion masks, clearer boundaries, and fewer background false responses, especially for small, low-contrast, and boundary-ambiguous targets.

Fig. 4(b)–(d) illustrates the effect of Dispersive Loss. Without dispersion regularization, the feature distributions tend to be more compact, and the response maps often leak into irrelevant background regions. When Dispersive Loss is applied, the features become more separable, the training curve becomes smoother, and the activation maps align more closely with the target regions. This supports its role in improving localization and boundary discrimination.

Fig. 5(d) shows that the attention-enhanced encoder–decoder design further improves intermediate responses.

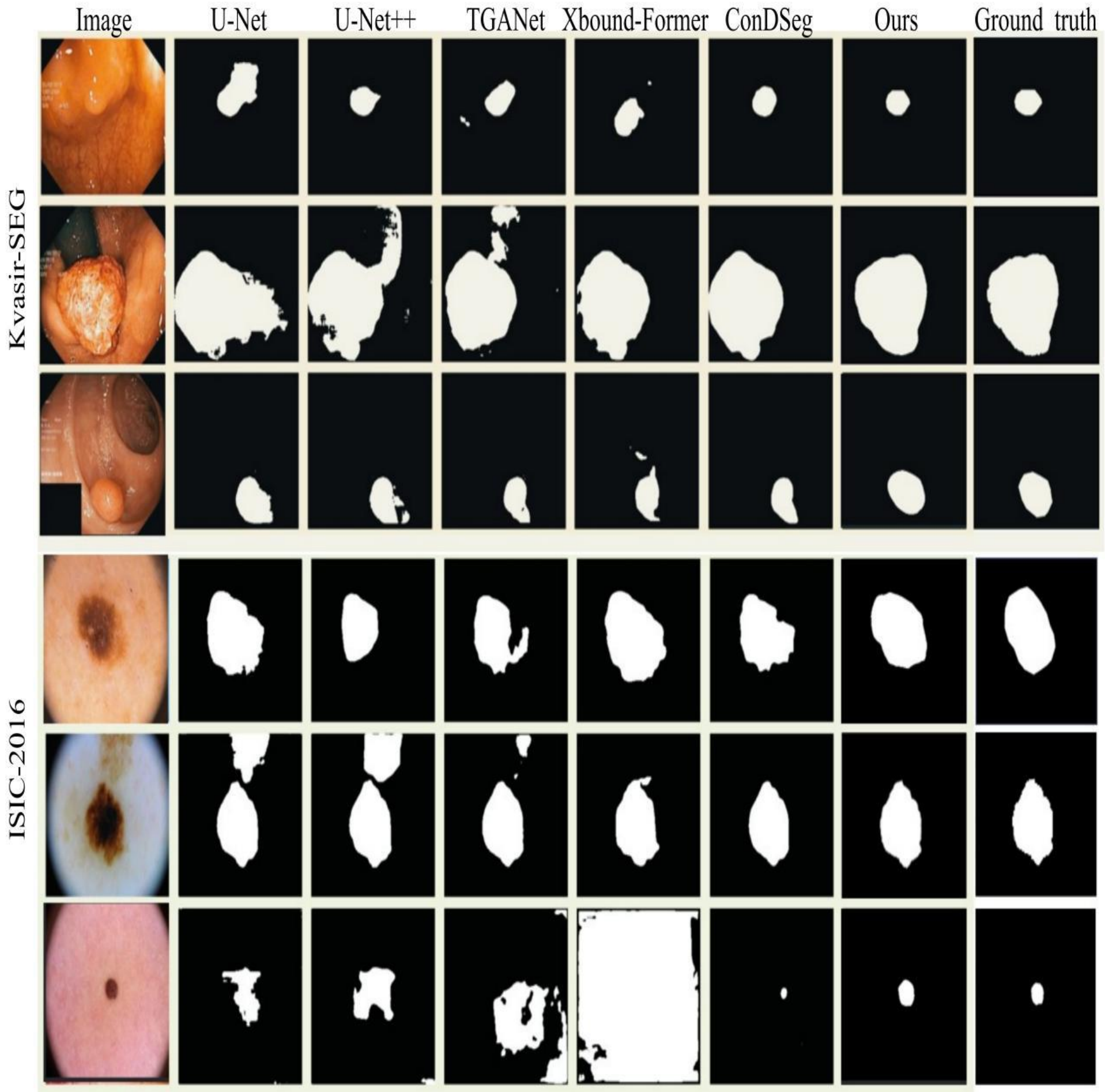

**Figure 6.** Visualisation of segmentation results of different models on the Kvasir-SEG and ISIC-2016 datasets.

Encoder-side attention suppresses distracting textures and highlights target-relevant regions, while multi-scale decoding provides complementary detail and contextual cues for more coherent mask reconstruction.

Fig. 7 presents a qualitative comparison with SAM 3 (Carion et al. 2025). Because SAM 3 is a general-purpose segmentation model with different input requirements, this comparison serves only as a qualitative reference rather than a strictly controlled benchmark. Under low contrast, flat lesions, or ambiguous boundaries, SAM 3 tends to produce fragmented or incomplete masks, while Med-DisSeg offers more stable lesion coverage and cleaner boundaries. This highlights the advantage of task-specific representation learning for medical image segmentation.

**Table 6.** Number of learnable parameters for different models at an input size of $256 \times 256$.

| Method | #Params (M) |
|---|---|
| TGANet | 19.84 |
| XBound-Former | 17.24 |
| ConDSeg | 45.55 |
| Ours (Med-DisSeg) | 45.56 |

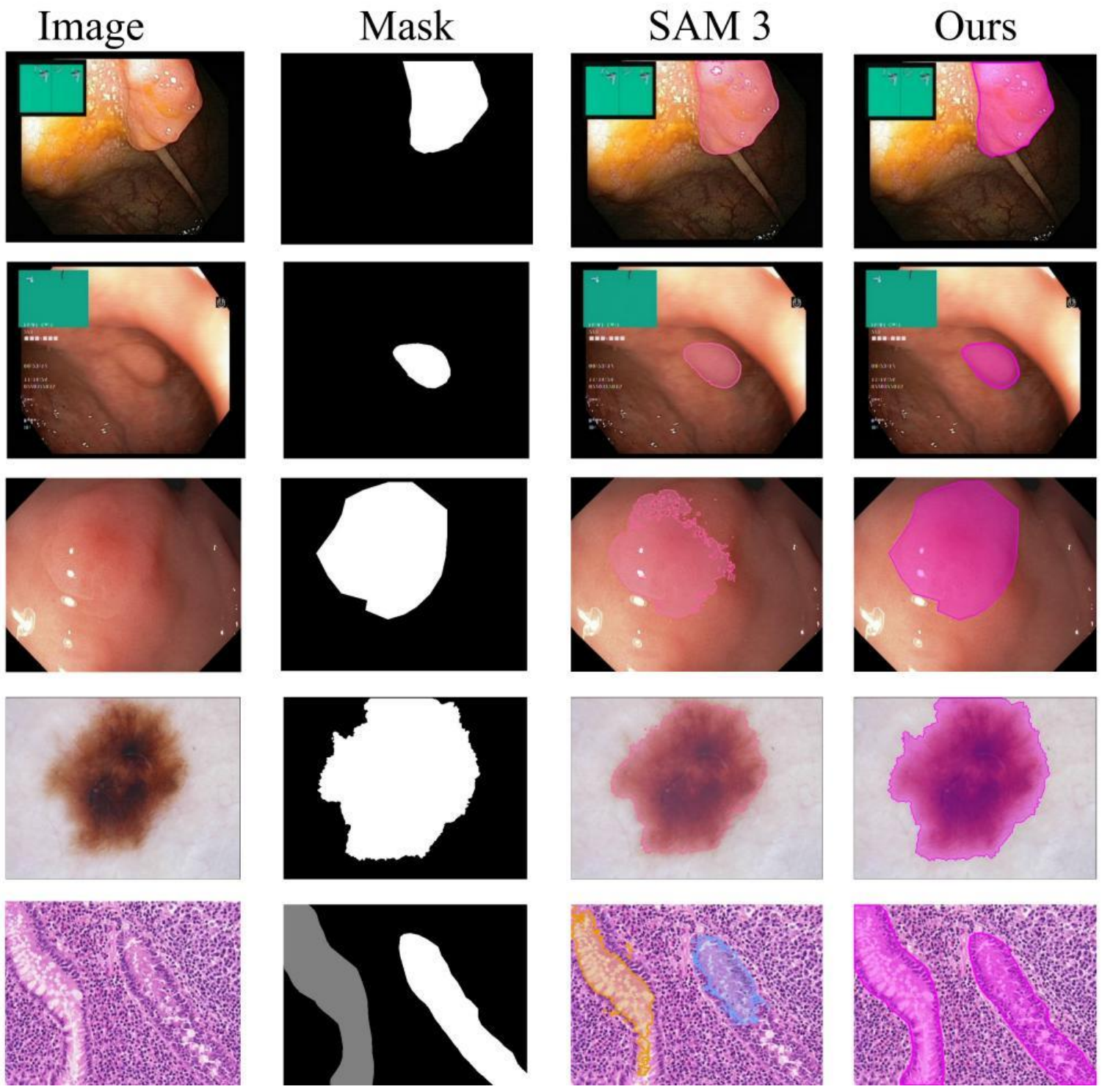

**Figure 7.** Visual comparison with the SAM 3 model.

**Table 7.** Inference speed on an NVIDIA GeForce RTX 4090 (fps).

| Method | TGANet | XBound-Former | ConDSeg | Ours |
|---|---|---|---|---|
| FPS | 32.7 | 61.3 | 59.6 | 59.6 |

## Computational Analysis

Table 6 compares the computational cost of Med-DisSeg with representative baselines at an input size of $256 \times 256$. Although Med-DisSeg has a parameter count comparable to ConDSeg (45.56M vs. 45.55M), it delivers consistently better segmentation performance across five benchmarks. This suggests that the gains stem primarily from the dispersion regularization and attention-enhanced multi-scale decoding, rather than from increased model capacity. Regarding efficiency, as shown in table 7， all FPS values are measured on a single NVIDIA RTX 4090 with a batch size of 4, excluding data loading and disk I/O. Under these settings, Med-DisSeg reaches 59.6 FPS, demonstrating practical inference speed.

## Conclusion

This paper proposed **Med-DisSeg**, a dispersion-driven framework for fine-grained medical image segmentation. By adopting Dispersive Loss as an auxiliary regularizer, Med-DisSeg learns more dispersed and separable feature representations. Building on these regularized representations, the encoder strengthens structure-relevant responses, while the decoder performs multi-scale calibration to improve the reconstruction of local details and boundaries. Experimental results on five public medical image segmentation datasets demonstrate the effectiveness and robustness of Med-DisSeg across diverse segmentation tasks.

## Acknowledgments

This work was supported by the National Key R&D Program of China (2023YFE0114300), the National Natural Science Foundation of China (U24A20652), and the Guangdong Basic and Applied Basic Research Foundation (2023A1515140146).

## Ethics Statement

This study has obtained the formal ethical approval from the Ethics Review Committee of Guangdong Provincial People's Hospital, with the ethical approval number KY2026-1124-01. All research procedures strictly comply with the relevant ethical guidelines for human-related research, and all participants provided informed consent before the study.